# Towards History-based Grammars:
# Using Richer Models for Probabilistic Parsing[*]

Ezra Black    Fred Jelinek    John Lafferty    David M. Magerman
Robert Mercer    Salim Roukos

IBM T. J. Watson Research Center

## ABSTRACT

We describe a generative probabilistic model of natural language, which we call HBG, that takes advantage of detailed linguistic information to resolve ambiguity. HBG incorporates lexical, syntactic, semantic, and structural information from the parse tree into the disambiguation process in a novel way. We use a corpus of bracketed sentences, called a Treebank, in combination with decision tree building to tease out the relevant aspects of a parse tree that will determine the correct parse of a sentence. This stands in contrast to the usual approach of further grammar tailoring via the usual linguistic introspection in the hope of generating the correct parse. In head-to-head tests against one of the best existing robust probabilistic parsing models, which we call P-CFG, the HBG model significantly outperforms P-CFG, increasing the parsing accuracy rate from 60% to 75%, a 37% reduction in error.

## 1. Introduction

Almost any natural language sentence is ambiguous in structure, reference, or nuance of meaning. Humans overcome these apparent ambiguities by examining the *context* of the sentence. But what exactly *is* context? Frequently, the correct interpretation is apparent from the words or constituents immediately surrounding the phrase in question. This observation begs the following question: How much information about the context of a sentence or phrase is necessary and sufficient to determine its meaning? This question is at the crux of the debate among computational linguists about the application and implementation of statistical methods in natural language understanding.

Previous work on disambiguation and probabilistic parsing has offered partial answers to this question. Hidden Markov models of words and their tags, introduced in [1] and [11] and popularized in the natural language community by Church [5], demonstrate the power of short-term $n$-gram statistics to deal with lexical ambiguity. Hindle and Rooth [8] use a statistical measure of lexical associations to resolve structural ambiguities. Brent [2] acquires likely verb subcategorization patterns using the frequencies of verb-object-preposition triples. Magerman and Marcus [10] propose a model of context that combines the $n$-gram model with information from dominating constituents. All of these aspects of context are necessary for disambiguation, yet none is sufficient.

We propose a probabilistic model of context for disambiguation in parsing, HBG, which incorporates the intuitions of these previous works into one unified framework. Let $p(T, w_1^n)$ be the joint probability of generating the word string $w_1^n$ and the parse tree $T$. Given $w_1^n$, our parser chooses as its parse tree that tree $T^*$ for which

$$T^* = \arg\max_{T \in \mathcal{P}(w_1^n)} p(T, w_1^n) \qquad (1)$$

where $\mathcal{P}(w_1^n)$ is the set of all parses produced by the grammar for the sentence $w_1^n$. Many aspects of the input sentence that might be relevant to the decision-making process participate in the probabilistic model, providing a very rich if not the richest model of context ever attempted in a probabilistic parsing model.

In this paper, we will motivate and define the HBG model, describe the task domain, give an overview of the grammar, describe the proposed HBG model, and present the results of experiments comparing HBG with an existing state-of-the-art model.

## 2. Motivation for History-based Grammars

One goal of a parser is to produce a grammatical interpretation of a sentence which represents the syntactic and semantic intent of the sentence. To achieve this goal, the parser must have a mechanism for estimating the coherence of an interpretation, both in isolation and in context. Probabilistic language models provide such a mechanism.

A probabilistic language model attempts to estimate the probability of a sequence of sentences and their respective interpretations (parse trees) occurring in the language, $\mathcal{P}(S_1\ T_1\ S_2\ T_2\ \ldots\ S_n\ T_n)$.

The difficulty in applying probabilistic models to natu-ral language is deciding what aspects of the sentence and the discourse are relevant to the model. Most previous probabilistic models of parsing assume the probabilities of sentences in a discourse are independent of other sentences. In fact, previous works have made much stronger independence assumptions. The P-CFG model considers the probability of each constituent rule independent of all other constituents in the sentence. The $\mathcal{P}$earl [10] model includes a slightly richer model of context, allowing the probability of a constituent rule to depend upon the immediate parent of the rule and a part-of-speech trigram from the input sentence. But none of these models come close to incorporating enough context to disambiguate many cases of ambiguity.

[*]Thanks to Philip Resnik and Stanley Chen for their valued input.

A significant reason researchers have limited the contextual information used by their models is because of the difficulty in estimating very rich probabilistic models of context. In this work, we present a model, the history-based grammar model, which incorporates a very rich model of context, and we describe a technique for estimating the parameters for this model using decision trees. The history-based grammar model provides a mechanism for taking advantage of contextual information from anywhere in the discourse history. Using decision tree technology, any question which can be asked of the history (i.e. Is the subject of the previous sentence animate? Was the previous sentence a question? etc.) can be incorporated into the language model.

## 3. The History-based Grammar Model

The history-based grammar model defines context of a parse tree in terms of the leftmost derivation of the tree.

Following [7], we show in Figure 1 a context-free grammar (CFG) for $a^n b^n$ and the parse tree for the sentence $aabb$. The leftmost derivation of the tree $T$ in Figure 1 is:

$$S \xrightarrow{r_1} ASB \xrightarrow{r_2} aSB \xrightarrow{r_3} aABB \xrightarrow{r_4} aaBB \xrightarrow{r_5} aabB \xrightarrow{r_6} aabb \quad (2)$$

where the rule used to expand the $i$-th node of the tree is denoted by $r_i$. Note that we have indexed the non-terminal (NT) nodes of the tree with this leftmost order. We denote by $t_i^-$ the sentential form obtained just before we expand node $i$. Hence, $t_3^-$ corresponds to the sentential form $aSB$ or equivalently to the string $r_1 r_2$. In a leftmost derivation we produce the words in left-to-right order.

Using the one-to-one correspondence between leftmost derivations and parse trees, we can rewrite the joint

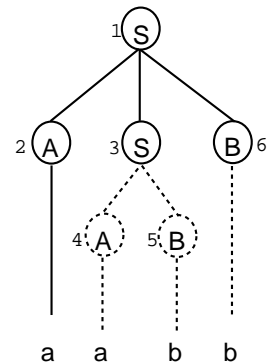

Figure 1: Grammar and parse tree for $aabb$.

probability in (1) as:

$$p(T, w_1^n) = \prod_{i=1}^{m} p(r_i | t_i^-)$$

In a probabilistic context-free grammar (P-CFG), the probability of an expansion at node $i$ depends only on the identity of the non-terminal $N_i$, i.e., $p(r_i | t_i^-) = p(r_i)$. Thus

$$p(T, w_1^n) = \prod_{i=1}^{m} p(r_i)$$

So in P-CFG the derivation order does not affect the probabilistic model[1].

A less crude approximation than the usual P-CFG is to use a decision tree to determine which aspects of the leftmost derivation have a bearing on the probability of how node $i$ will be expanded. In other words, the probability distribution $p(r_i | t_i^-)$ will be modeled by $p(r_i | E[t_i^-])$ where $E[t]$ is the equivalence class of the history $t$ as determined by the decision tree. This allows our probabilistic model to use any information anywhere in the partial derivation tree to determine the probability of different expansions of the $i$-th non-terminal. The use of decision trees and a large bracketed corpus may shift some of the burden of identifying the intended parse from the grammarian to the statistical estimation methods. We refer to probabilistic methods based on the derivation as History-based Grammars (HBG).

---

[1]Note the abuse of notation since we denote by $p(r_i)$ the conditional probability of rewriting the non-terminal $N_i$.

In this paper, we explored a restricted implementation of this model in which only the path from the current node to the root of the derivation along with the index of a branch (index of the child of a parent ) are examined in the decision tree model to build equivalence classes of histories. Other parts of the subtree are not examined in the implementation of HBG.

## 4. Task Domain

We have chosen computer manuals as a task domain. We picked the most frequent 3000 words in a corpus of 600,000 words from 10 manuals as our vocabulary. We then extracted a few million words of sentences that are completely covered by this vocabulary from 40,000,000 words of computer manuals. A randomly chosen sentence from a sample of 5000 sentences from this corpus is:

396.   It indicates whether a call completed successfully or if some error was detected that caused the call to fail.

To define what we mean by a correct parse, we use a corpus of manually bracketed sentences at the University of Lancaster called the Treebank. The Treebank uses 17 non-terminal labels and 240 tags. The bracketing of the above sentence is shown in Figure 2.

```
[N It_PPH1 N]
[V indicates_VVZ
   [Fn [Fn&whether_CSW
          [N a_AT1 call_NN1 N]
          [V completed_VVD successfully_RR V]Fn&]
       or_CC
   [Fn+ if_CSW
          [N some_DD error_NN1 N]@
          [V was_VBDZ detected_VVN V]
          @[Fr that_CST
             [V caused_VVD
                [N the_AT call_NN1 N]
                [Ti to_TO fail_VVI Ti]V]Fr]Fn+]
                                    Fn]V]._
```

Figure 2: Sample bracketed sentence from Lancaster Treebank.

A parse produced by the grammar is judged to be correct if it agrees with the Treebank parse structurally and the NT labels agree. The grammar has a significantly richer NT label set (more than 10000) than the Treebank but we have defined an equivalence mapping between the grammar NT labels and the Treebank NT labels. In this paper, we do not include the tags in the measure of a correct parse.

We have used about 25,000 sentences to help the grammarian develop the grammar with the goal that the correct (as defined above) parse is among the proposed (by the grammar) parses for a sentence. Our most common test set consists of 1600 sentences that are never seen by the grammarian.

## 5. The Grammar

The grammar used in this experiment is a broad-coverage, feature-based unification grammar. The grammar is context-free but uses unification to express rule templates for the the context-free productions. For example, the rule template:

$$\left[ \begin{array}{c} NP \\ :n \end{array} \right] \rightarrow \left[ \begin{array}{c} Det \\ unspec \end{array} \right] \left[ \begin{array}{c} N \\ :n \end{array} \right] \qquad (3)$$

corresponds to three CFG productions where the second feature $:n$ is either $s$, $p$, or $:n$. This rule template may elicit up to 7 non-terminals. The grammar has 21 features whose range of values maybe from 2 to about 100 with a median of 8. There are 672 rule templates of which 400 are actually exercised when we parse a corpus of 15,000 sentences. The number of productions that are realized in this training corpus is several hundred thousand.

### 5.1. P-CFG

While a NT in the above grammar is a feature vector, we group several NTs into one class we call a **mnemonic** represented by the one NT that is the least specified in that class. For example, the mnemonic VB0PASTSG* corresponds to all NTs that unify with:

$$\left[ \begin{array}{c} pos = v \\ v - type = be \\ tense - aspect = past \end{array} \right] \qquad (4)$$

We use these mnemonics to label a parse tree and we also use them to estimate a P-CFG, where the probability of rewriting a NT is given by the probability of rewriting the mnemonic. So from a training set we induce a CFG from the actual mnemonic productions that are elicited in parsing the training corpus. Using the Inside-Outside algorithm, we can estimate P-CFG from a large corpus of text. But since we also have a large corpus of bracketed sentences, we can adapt the Inside-Outside algorithm to reestimate the probability parameters subject to the constraint that only parses consistent with the Treebank (where consistency is as defined earlier) contribute to the reestimation. From a training run of 15,000 sentences we observed 87,704 mnemonic productions, with 23,341 NT mnemonics of which 10,302 were lexical. Running on a test set of 760 sentences 32% of the rule templates were used, 7% of the lexical mnemonics, 10% of the constituent mnemonics, and 5% of the mnemonic productions actually contributed to parses of test sentences.

## 5.2. Grammar and Model Performance Metrics

To evaluate the performance of a grammar and an accompanying model, we use two types of measurements:

- the *any-consistent* rate, defined as the percentage of sentences for which the correct parse is proposed among the many parses that the grammar provides for a sentence. We also measure the *parse base*, which is defined as the geometric mean of the number of proposed parses on a per word basis, to quantify the ambiguity of the grammar.

- the *Viterbi* rate defined as the percentage of sentences for which the most likely parse is consistent.

The *any-consistent* rate is a measure of the grammar's coverage of linguistic phenomena. The *Viterbi* rate evaluates the grammar's coverage with the statistical model imposed on the grammar. The goal of probabilistic modelling is to produce a *Viterbi* rate close to the *any-consistent* rate.

The *any-consistent* rate is 90% when we require the structure and the labels to agree and 96% when unlabeled bracketing is required. These results are obtained on 760 sentences from 7 to 17 words long from test material that has never been seen by the grammarian. The *parse base* is 1.35 parses/word. This translates to about 23 parses for a 12-word sentence. The unlabeled *Viterbi* rate stands at 64% and the labeled *Viterbi* rate is 60%.

While we believe that the above *Viterbi* rate is close if not the state-of-the-art performance, there is room for improvement by using a more refined statistical model to achieve the labeled *any-consistent* rate of 90% with this grammar. There is a significant gap between the labeled *Viterbi* and *any-consistent* rates: 30 percentage points.

Instead of the usual approach where a grammarian tries to fine tune the grammar in the hope of improving the *Viterbi* rate we use the combination of a large Treebank and the resulting derivation histories with a decision tree building algorithm to extract statistical parameters that would improve the *Viterbi* rate. The grammarian's task remains that of improving the *any-consistent* rate.

The history-based grammar model is distinguished from the context-free grammar model in that each constituent structure depends not only on the input string, but also the entire history up to that point in the sentence. In HBGs, history is interpreted as any element of the output structure, or the parse tree, which has already been determined, including previous words, non-terminal categories, constituent structure, and any other linguistic information which is generated as part of the parse structure.

## 6. The HBG Model

Unlike P-CFG which assigns a probability to a mnemonic production, the HBG model assigns a probability to a rule template. Because of this the HBG formulation allows one to handle any grammar formalism that has a derivation process.

For the HBG model, we have defined about 50 syntactic categories, referred to as $Syn$, and about 50 semantic categories, referred to as $Sem$. Each NT (and therefore mnemonic) of the grammar has been assigned a syntactic ($Syn$) and a semantic ($Sem$) category. We also associate with a non-terminal a primary lexical head, denoted by $H_1$, and a secondary lexical head, denoted by $H_2$.[2] When a rule is applied to a non-terminal, it indicates which child will generate the lexical primary head and which child will generate the secondary lexical head.

The proposed generative model associates for each constituent in the parse tree the probability:

$p(Syn, Sem, R, H_1, H_2 \mid Syn_p, Sem_p, R_p, I_{pc}, H_{1p}, H_{2p})$

In HBG, we predict the syntactic and semantic labels of a constituent, its rewrite rule, and its two lexical heads using the labels of the parent constituent, the parent's lexical heads, the parent's rule $R_p$ that lead to the constituent and the constituent's index $I_pc$ as a child of $R_p$. As we discuss in a later section, we have also used with success more information about the derivation tree than the immediate parent in conditioning the probability of expanding a constituent.

We have approximated the above probability by the following five factors:

1. $p(Syn \mid R_p, I_{pc}, H_{1p}, Syn_p, Sem_p)$

---

[2] The primary lexical head $H_1$ corresponds (roughly) to the linguistic notion of a lexical head. The secondary lexical head $H_2$ has no linguistic parallel. It merely represents a word in the constituent besides the head which contains predictive information about the constituent.

2. $p(Sem \mid Syn, R_p, I_{pc}, H_{1p}, H_{2p}, Syn_p, Sem_p)$

3. $p(R \mid Syn, Sem, R_p, I_{pc}, H_{1p}, H_{2p}, Syn_p, Sem_p)$

4. $p(H_1 \mid R, Syn, Sem, R_p, I_{pc}, H_{1p}, H_{2p})$

5. $p(H_2 \mid H_1, R, Syn, Sem, R_p, I_{pc}, Syn_p)$

While a different order for these predictions is possible, we only experimented with this one.

### 6.1. Parameter Estimation

We only have built a decision tree to the rule probability component (3) of the model. For the moment, we are using *n-gram* models with the usual *deleted interpolation* for smoothing for the other four components of the model.

We have assigned bit strings to the syntactic and semantic categories and to the rules manually. Our intention is that bit strings differing in the least significant bit positions correspond to categories of non-terminals or rules that are similar. We also have assigned bitstrings for the words in the vocabulary (the lexical heads) using automatic clustering algorithms using the bigram mutual information clustering algorithm (see [4]). Given the bitsting of a history, we then designed a decision tree for modeling the probability that a rule will be used for rewriting a node in the parse tree.

Since the grammar produces parses which may be more detailed than the Treebank, the decision tree was built using a training set constructed in the following manner. Using the grammar with the P-CFG model we determined the most likely parse that is consistent with the Treebank and considered the resulting sentence-tree pair as an event. Note that the grammar parse will also provide the lexical head structure of the parse. Then, we extracted using leftmost derivation order tuples of a history (truncated to the definition of a history in the HBG model) and the corresponding rule used in expanding a node. Using the resulting data set we built a decision tree by classifying histories to locally minimize the entropy of the rule template.

With a training set of about 9000 sentence-tree pairs, we had about 240,000 tuples and we grew a tree with about 40,000 nodes. This required 18 hours on a 25 MIPS RISC-based machine and the resulting decision tree was nearly 100 megabytes.

### 6.2. Immediate vs. Functional Parents

The HBG model employs two types of parents, the *immediate* parent and the *functional* parent. The immediate parent is the constituent that immediately dominates the constituent being predicted. If the immediate parent of a constituent has a different syntactic type from that of the constituent, then the immediate parent is also the functional parent; otherwise, the functional parent is the functional parent of the immediate parent. The distinction between functional parents and immediate parents arises primarily to cope with unit productions. When unit productions of the form XP2 → XP1 occur, the immediate parent of XP1 is XP2. But, in general, the constituent XP2 does not contain enough useful information for ambiguity resolution. In particular, when considering only immediate parents, unit rules such as NP2 → NP1 prevent the probabilistic model from allowing the NP1 constituent to interact with the VP rule which is the functional parent of NP1.

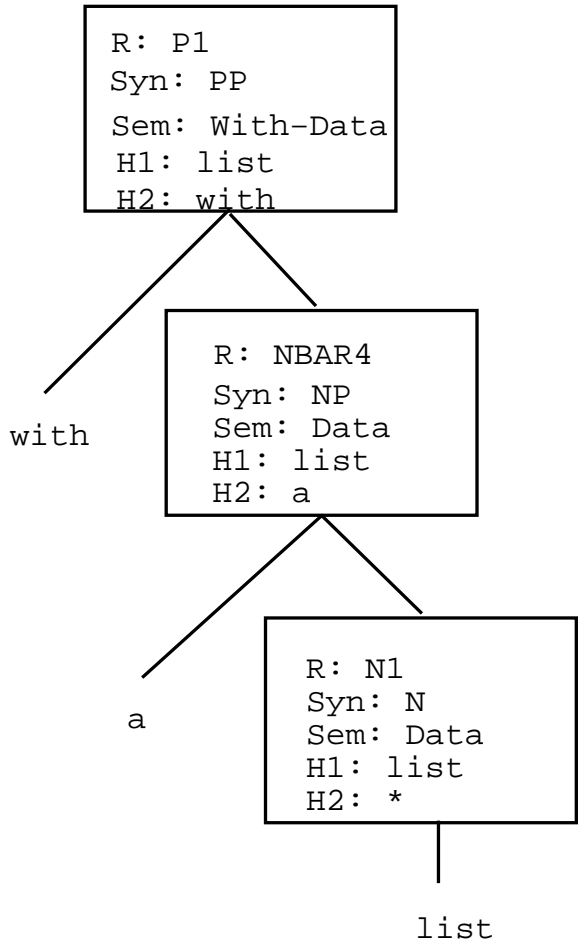

Figure 3: Sample representation of "with a list" in HBG model.

When the two parents are identical as it often happens, the duplicate information will be ignored. However, when they differ, the decision tree will select that parental context which best resolves ambiguities.

Figure 3 shows an example of the representation of a history in HBG for the prepositional phrase "with a list." In this example, the immediate parent of the N1 node is the NBAR4 node and the functional parent of N1 is the PP1 node.

## 7. Results

We compared the performance of HBG to the "broad-coverage" probabilistic context-free grammar, P-CFG. The *any-consistent* rate of the grammar is 90% on test sentences of 7 to 17 words. The *Viterbi* rate of P-CFG is 60% on the same test corpus of 760 sentences used in our experiments. On the same test sentences, the HBG model has a *Viterbi* rate of 75%. This is a reduction of 37% in error rate.

|  | Accuracy |
| --- | --- |
| P-CFG | 59.8% |
| HBG | 74.6% |
| Error Reduction | 36.8% |

Figure 4: Parsing accuracy: P-CFG vs. HBG

In developing HBG, we experimented with similar models of varying complexity. One discovery made during this experimentation is that models which incorporated more context than HBG performed slightly worse than HBG. This suggests that the current training corpus may not contain enough sentences to estimate richer models. Based on the results of these experiments, it appears likely that significantly increasing the size of the training corpus should result in a corresponding improvement in the accuracy of HBG and richer HBG-like models.

To check the value of the above detailed history, we tried the simpler model:

1. $p(H_1 \,|\, H_{1p}, H_{2p}, R_p, I_{pc})$
2. $p(H_2 \,|\, H_1, H_{1p}, H_{2p}, R_p, I_{pc})$
3. $p(Syn \,|\, H_1, R_p, I_{pc})$
4. $p(Sem \,|\, Syn, H_1, R_p, I_{pc})$
5. $p(R \,|\, Syn, Sem, H_1, H_2)$

This model corresponds to a P-CFG with NTs that are the crude syntax and semantic categories annotated with the lexical heads. The *Viterbi* rate in this case was 66%, a small improvement over the P-CFG model indicating the value of using more context from the derivation tree.

## 8. Conclusions

The success of the HBG model encourages future development of general history-based grammars as a more promising approach than the usual P-CFG. More experimentation is needed with a larger Treebank than was used in this study and with different aspects of the derivation history. In addition, this paper illustrates a new approach to grammar development where the parsing problem is divided (and hopefully conquered) into two subproblems: one of grammar coverage for the grammarian to address and the other of statistical modeling to increase the probability of picking the correct parse of a sentence.